\documentclass[11pt,a4paper]{article}
\usepackage[hyperref]{acl2020}
\usepackage{times}
\usepackage{latexsym}

\usepackage[belowskip=-6pt,aboveskip=2pt]{caption}

\usepackage[utf8]{inputenc}
\usepackage{bbm}

\usepackage{xcolor}
\usepackage{soul}
\usepackage[utf8]{inputenc}
\usepackage{amsmath}
\usepackage{amssymb}

\usepackage{color}
\usepackage{xspace} 

\usepackage{times}
\usepackage{helvet}
\usepackage{courier}
\usepackage{pgfplots}
\usepackage{graphicx}
\usepackage{subfig}
\usepackage{subfloat}
\usepackage{tikz-dependency}

\usepackage{algorithm}
\usepackage{algorithmic}

\usepackage{tabularx}
\usepackage{subfig}
\usepackage{xparse}
\usepackage{xspace}
\usepackage{relsize}
\usepackage{graphicx}
\usepackage{multirow}
\usepackage[normalem]{ulem}
\usepackage{amsmath}
\usepackage{bm}
\usepackage{url}
\usepackage{booktabs}
\usepackage{xcolor}
\usepackage{multirow}
\usepackage{colortbl}

\usepackage{times}  
\usepackage{helvet}  
\usepackage{courier}  
\usepackage{url}  
\usepackage{graphicx}  

\usepackage{float}
\usepackage{graphicx}
\usepackage{placeins}

\usepackage{amssymb}
\usepackage{breqn}

\def\x{\mathbf{x}}
\def\y{\mathbf{y}}
\def\bu{\mathbf{u}}

\def\bz{\mathbf{z}}

\def\h{\mathbf{h}}
\def\e{\mathbf{e}}

\def\g{\mathbf{g}}

\allowdisplaybreaks

\DeclareMathOperator*{\LSTM}{LSTM}
\DeclareMathOperator*{\ATT}{ATT}
\DeclareMathOperator*{\softmax}{softmax}

\makeatletter

\newcommand{\Rmnum}[1]{\expandafter\@slowromancap\romannumeral #1@}

\aclfinalcopy 

\pgfplotsset{compat=1.7}
\pgfplotsset{every axis/.append style={
                    axis x line=middle,    
                    axis y line=middle,    
                    axis line style={->}, 
                    xlabel={$x$},          
                    ylabel={$y$},          
                    label style={font=\tiny},
                    tick label style={font=\tiny},
                    legend style={font=\tiny},
                    legend pos=outer north east
                    }}

\tikzset{>=stealth}

\usepackage{microtype}

\def\aclpaperid{2176} 

\title{Improving Image Captioning with Better Use of Captions}

\author{Zhan Shi$^{*}$, Xu Zhou$^{\dagger}$, Xipeng Qiu$^{\dagger}$, Xiaodan Zhu$^{*}$ \\
  $^{*}$Ingenuity Labs Research Institute, Queen’s University \\
  $^{*}$Department of Electrical and Computer Engineering, Queen’s University \\
  $^{\dagger}$School of Computer Science, Fudan University\\
  \texttt{\{z.shi,xiaodan.zhu\}@queensu.ca,\{16210240095,xpqiu\}@fudan.edu.cn} \\}

\date{}

\begin{document}
\maketitle
\begin{abstract}
Image captioning is a multimodal problem that has drawn extensive attention in both the natural language processing and computer vision community. In this paper, we present a novel image captioning architecture to better explore semantics available in captions and leverage that to enhance both image representation and caption generation. Our models first construct caption-guided visual relationship graphs that introduce beneficial inductive bias using weakly supervised multi-instance learning. The representation is then enhanced with neighbouring and contextual nodes with their textual and visual features. During generation, the model further incorporates visual relationships using multi-task learning for jointly predicting word and object/predicate tag sequences.
We perform extensive experiments on the MSCOCO dataset, showing that the proposed framework significantly outperforms the baselines, resulting in the state-of-the-art performance under a wide range of evaluation metrics. The code of our paper has been made publicly available.~\footnote{ https://github.com/Gitsamshi/WeakVRD-Captioning}
\end{abstract}

\section{Introduction}
Automatically generating a short description for a given image, a problem known as image captioning~\cite{chen2015microsoft}, has drawn extensive attention in both the natural language processing and computer vision community. 
Inspired by the success of encoder-decoder frameworks with the attention mechanism,  previous efforts on image captioning adopt variants of pre-trained convolution neural networks (CNN) as the image encoder and recurrent neural networks (RNN) with visual attention as the decoder~\cite{lu2017knowing,anderson2018bottom,xu2015show,lu2018neural}.

\begin{figure}[t]
  \centering
  \includegraphics[width=0.45\textwidth]{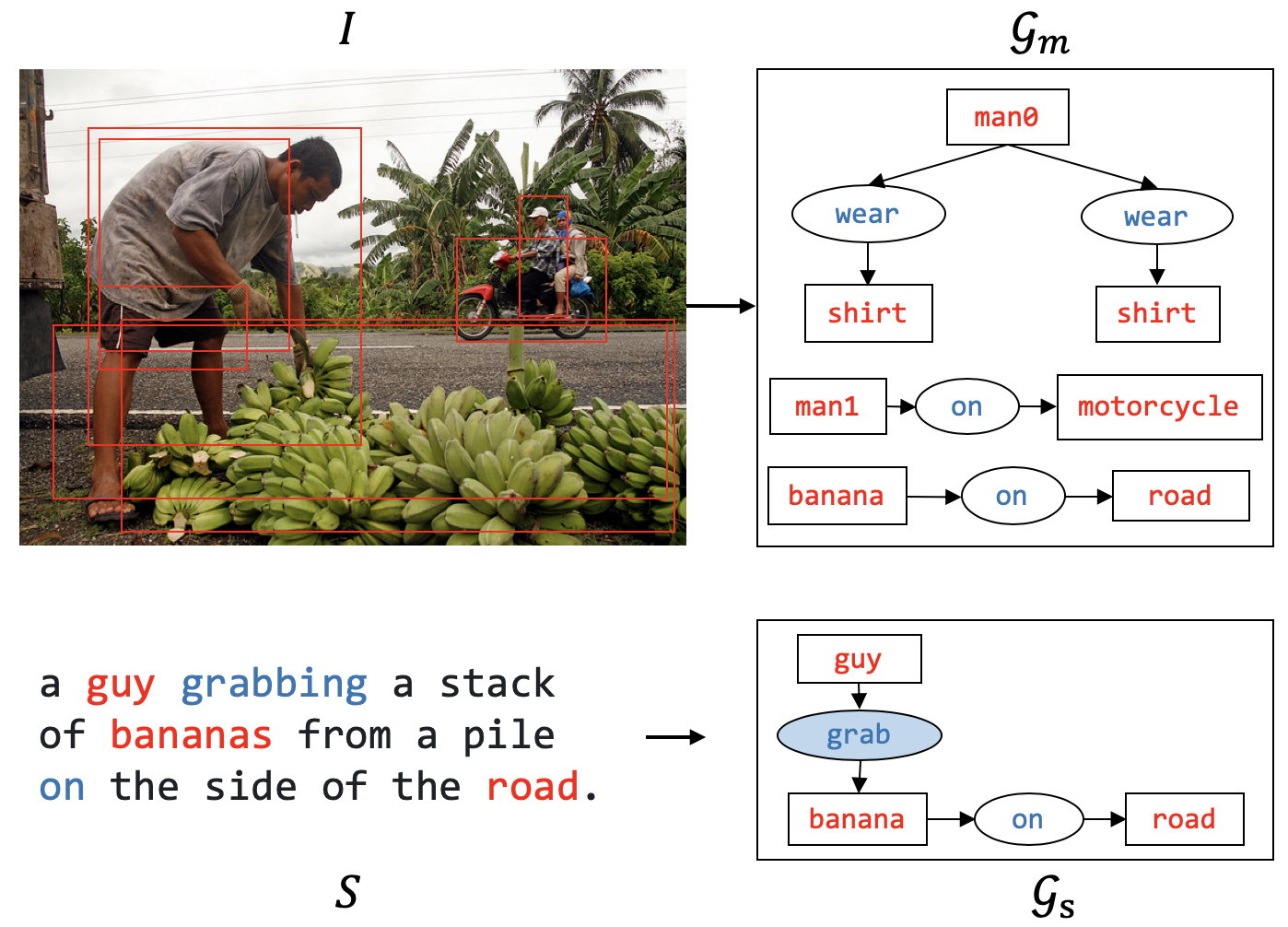}
  \caption{Visual relationship graphs from a pre-trained detection model~\cite{yao2018exploring} (upper) and from the ground-truth caption (bottom). }
  \label{fig:weakness}
\end{figure}

Many previous methods translate image representation into natural language sentences without explicitly investigating semantic cues from texts and images. To remedy that, some research has also explored to detect high-level semantic concepts presented in images to improve caption generation~\cite{wu2016value,gan2017semantic,you2016image,fang2015captions,yao2017boosting}. It is believed by many that the inductive bias that leverages structured combination of concepts and visual relationships is of importance, which has led to better captioning models~\cite{yao2018exploring,guo2019aligning,yang2019auto}. These approaches obtain visual relationship graphs using models pre-trained from visual relationship detection (VRD) datasets, e.g., Visual Genome~\cite{krishna2017visual}, where the visual relationships capture semantics between pairs of localized \textit{objects} connected by \textit{predicates}, including spatial (e.g., \textit{cake-on-desk}) and non-spatial semantic relationships (e.g., \textit{man-eat-food})~\cite{lu2016visual}.

As in many other joint text-image modeling problems, it is crucial to obtain a good semantic representation in image captioning that bridges semantics in language and images. The existing approaches, however, have not yet adequately leveraged the semantics available in captions to construct image representation and generate captions. As shown in Figure~\ref{fig:weakness}, although VRD detection models present a strong capacity in predicting salient objects and the most common predicates,
they often ignore predicates vital for captioning (e.g., ``grab" in this example). Exploring better models would still be highly desirable. 

A major challenge for establishing a structural
connection between captions and images is that the links between predicates and the corresponding object regions are often ambiguous:
within the ``image-level" label $(obj_1, pred, obj_2)$ extracted from captions, there may exist multiple object regions corresponding to $obj_1$ and $obj_2$. In this paper, we propose to use weakly supervised multi-instance learning to detect if a bag of object (region) pairs in an image contain certain predicates, e.g., predicates appearing in ground-truth captions here (or in other applications, they can be any given predicates under concerns). Based on that we can construct caption-guided visual relationship graphs.

Once the visual relationship graphs (VRG) are built, we propose to adapt graph convolution operations~\cite{marcheggiani2017encoding} to obtain representation for object nodes and predicate nodes. These nodes can be viewed as image representation units used for generation. 

During generation, we further incorporate visual relationships---we propose multi-task learning for jointly predicting word and tag sequences, where each word in a caption could be assigned with a tag, i.e., \textit{object}, \textit{predicate}, or \textit{none}, which takes as input the graph node features from the above visual relationship graphs. The motivation for predicting a tag in each step is to regularize which types of information should be taken into more consideration for generating words: predicate nodes features, object nodes features, or the current state of language decoder. We study different types of multi-task blocks in our models.  

As a result, our models consist of three major components: constructing caption-guided visual relationship graphs (CGVRG) with weakly-supervised multi-instance learning, building context-aware CGVRG, and performing multi-task generation to regularize the network to take into account explicit predicate object/predicate constraints. We perform extensive experiments on the MSCOCO~\cite{lin2014microsoft} image captioning dataset with both supervised and Reinforcement learning strategy~\cite{rennie2017self}. The experiment results show that the proposed models significantly outperform the baselines and achieve the state-of-the-art performance under a wide range of evaluation metrics. The main contributions of our work are summarized as follows:
\begin{itemize}
\setlength\itemsep{-0.2em}
    \item We propose to construct caption-guided visual relationship graphs that introduce beneficial inductive bias by better bridging captions and images. The representation is further enhanced with neighbouring and contextual nodes with their textual and visual features.
    \item Unlike existing models, we propose  multi-task learning to regularize the network to take into account explicit object/predicate constraints in the process of generation. 
    \item The proposed framework achieves the state-of-the-art performance on the MSCOCO image captioning dataset. We provide detailed analyses on how this is attained.
\end{itemize}

\section{Related Work}
\begin{figure*}[t]
  \centering
  \includegraphics[width=0.82\textwidth]{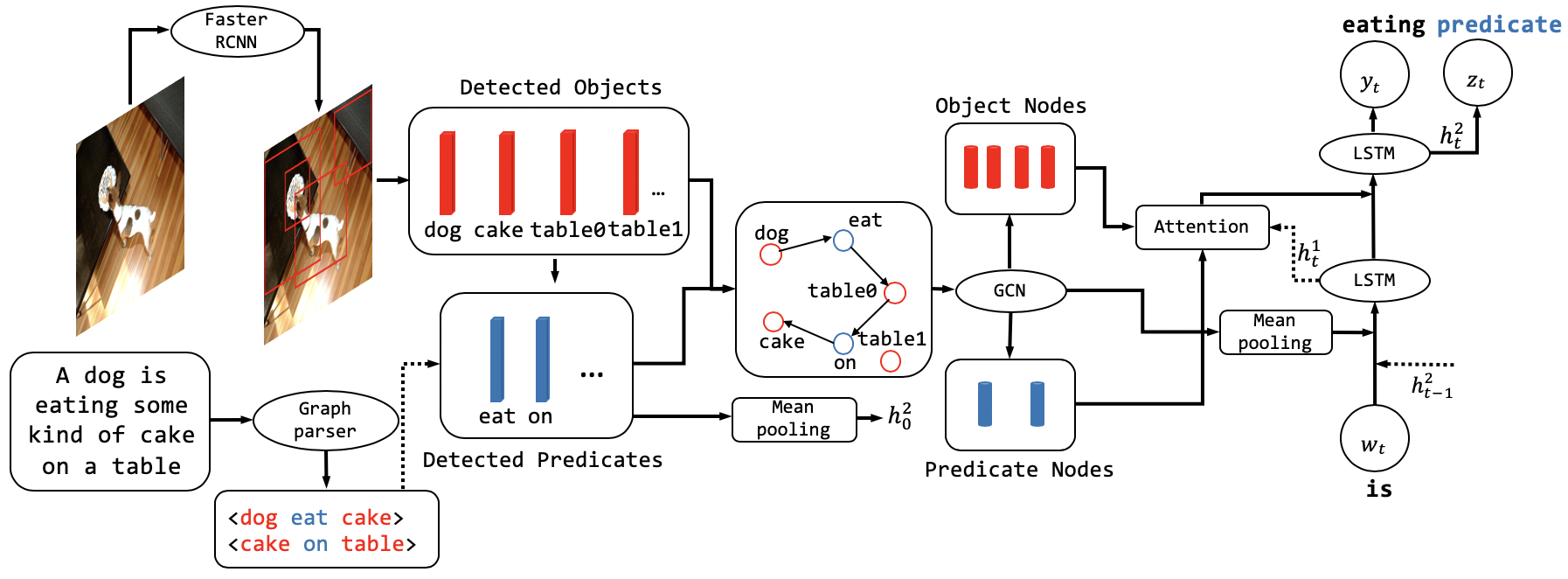}
  \caption{An overview of the proposed image captioning framework.}\label{fig:framework}
\end{figure*}
\textbf{Image Captioning} A prevalent paradigm of existing image captioning methods is based on the encoder-decoder framework which often utilizes a CNN-plus-RNN architecture for image encoding and text generation~\cite{donahue2015long,vinyals2015show,karpathy2015deep}. Soft or hard visual attention mechanism~\cite{xu2015show,chen2017sca} has been incorporated to focus on the most relevant regions in each generation step.
Furthermore, adaptive attention~\cite{lu2017knowing} has been developed to decide whether to rely on visual features or language model states in each decoding step. 
Recently, bottom-up attention techniques~\cite{anderson2018bottom,lu2018neural} have also been proposed to find the most relevant regions based on bounding boxes. 

There has been increasing work focusing on filling the gap between image representation and caption generation. Semantic concepts and attributes detected from images have been demonstrated to be effective in boosting image captioning when used in the encoder-decoder frameworks~\cite{wu2016value,you2016image,gan2017semantic,yao2017boosting}. Visual relationship~\cite{lu2016visual} and scene graphs~\cite{johnson2015image} have been further employed for image encoder in a uni-modal~\cite{yao2018exploring} or multi-modal~\cite{yang2019auto,guo2019aligning} manner to improve the overall performance via the graph convolutional mechanism~\cite{marcheggiani2017encoding}. Besides, \newcite{kim2019dense} proposes a relationship-based captioning task to lead better understanding of images based on relationship. As discussed in introduction, we will further explore the relational semantics available in captions for both constructing image representation and generating caption.

\paragraph{Visual Relationship Detection} Visual relations between objects in an image have attracted more studies recently. Conventional visual relation detection have
dealt with $\langle \textit{subject-predicate-object}\rangle$ triples, including spatial relation and other semantic relation.~\citet{lu2016visual} detect the triples
by performing subject, object, and predicate classification separately.~\citet{li2017vip} attempt to encode more distinguishable visual features for visual relationships detection. Probabilistic output of object detection~\cite{dai2017detecting,zhang2017visual} is also considered to reason about the visual relationships.

Given an image $\bm{I}$, the goal of
image captioning is to generate a visually grounded natural language sentence. We learn our model by minimizing the cross-entropy loss with regard to the ground truth caption $\bm{S}^{*} = \{w_1^*, w_2^*, ..., w_T^*\}$:

\begin{align}
    L_{XE} &= - \log p(\bm{S^*}|\bm{I}) \\
    & = - \sum_{t=1}^T \log p(w_t^*|w_{<t}^*, \bm{I})
\end{align}
The model is further tuned with a Reinforcement Learning (RL) objective~\cite{rennie2017self} to maximize the reward of the generated sentence $\bm{S}$:
\begin{equation}
    J_{RL} = E_{\bm{S} \sim p(\bm{S}|\bm{I})} (d(\bm{S},\bm{S}^*))
\label{eq:rl_loss}
\end{equation}
where $d$ is a sentence-level scoring metric.  

An overview of our image captioning framework is depicted in Figure~\ref{fig:framework}, with the detail of the components described in the following sections.

\subsection{Caption-Guided Visual Relationship Graph (CGVRG) with Weakly Supervised Learning}

A general challenge of modeling $p(\bm{S}|\bm{I})$ is obtaining a better semantic representation in the multimodal setting to bridge captions and images. Our framework first focuses on constructing caption-guided visual relationship graphs (CGVRG).


\subsubsection{Extracting Visual Relationship Triples and Detecting Objects} 
The process of constructing CGVRG first extracts relationship triples from captions using textual scene graph parser as described in \cite{schuster2015generating}. Our framework employs Faster R-CNN~\cite{ren2015faster} to recognize instances of objects and returns a set of image regions for objects: $V = \{v_1, v_2, \cdots, v_n\}$. 

\subsubsection{Constructing CGVRG} 
The main focus of CGVRG is constructing visual relationship graphs. As discussed in introduction, the existing approaches use pre-trained VRD (visual relationship detection) models, which often ignore key relationships needed for captioning. This gap can be even more prominent if the domain/data used to train image-captioning is farther from where VRD is pretrained. A major challenge to use predicate triples from captions to construct CGVRG is that, the links between predicates and the corresponding object regions are often ambiguous as discussed in introduction. To solve this problem, we use weakly supervised, multi-instance learning. 

\begin{figure}[t]
  \centering
  \includegraphics[width=0.45\textwidth]{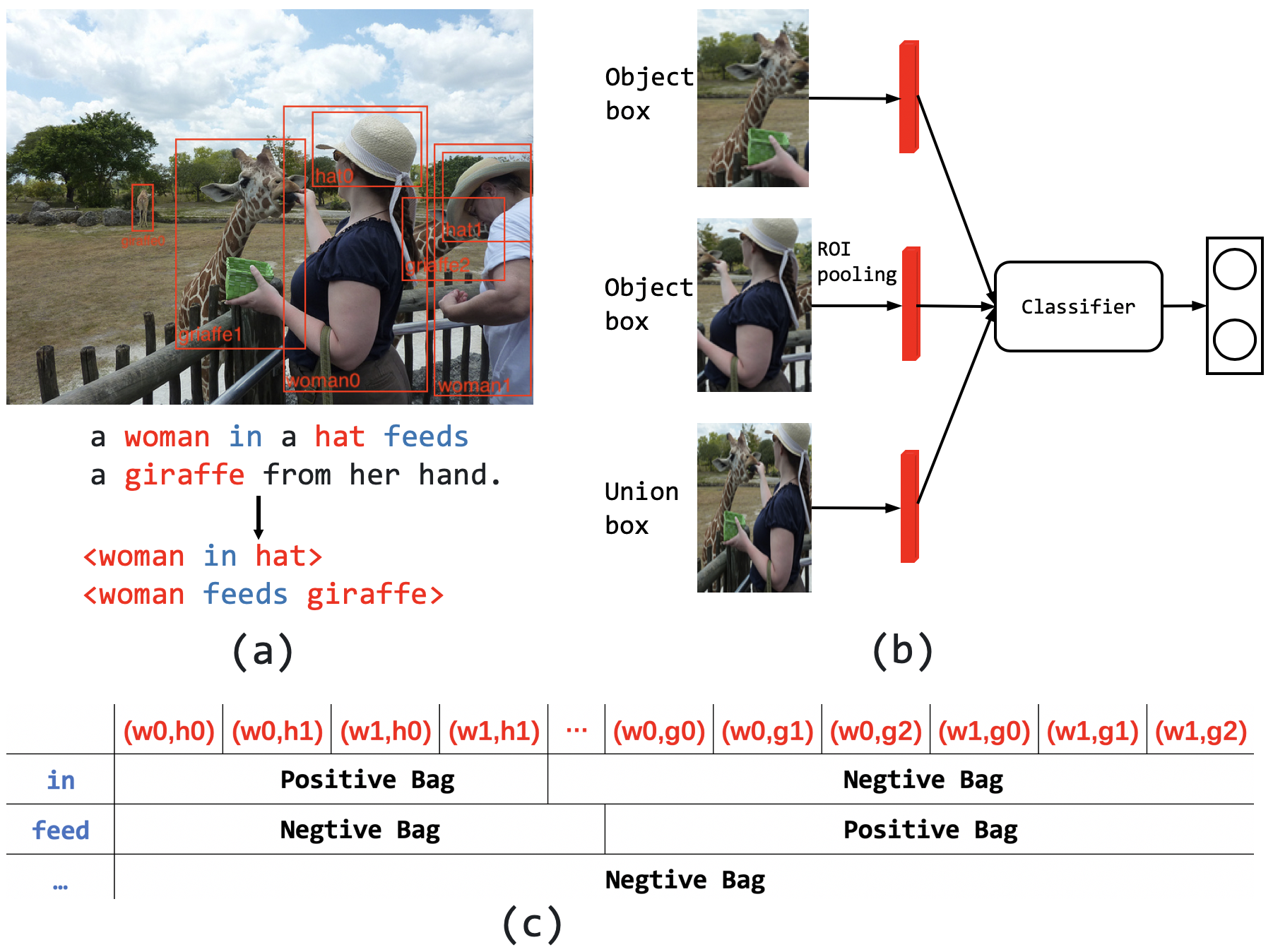}
  \caption{Subcomponents in constructing CGVRG: (a) detecting objects and extracting triples; (b) obtaining representation for object region pairs;  (c) examples of positive and negative bags in multi-instance learning for predicate ``in" and ``feed", respectively. Here, $w$, $h$, and $g$ denote \textit{woman}, \textit{hat}, and \textit{giraffe}, respectively.}\label{fig:vrd}
\end{figure}

\paragraph{Obtaining Representation for Object Region Pairs} For an image $I$ with a list of salient object regions  obtained in object detection $\{v_1, v_2, \cdots, v_n\}$, we have a set of region pairs $U = \{\bu_1, \bu_2, \cdots, \bu_N\}$, where $N=n(n-1)$.
As shown in Figure~\ref{fig:vrd}(b), the visual features of any two object regions and their union box will be collected to compute $p_{\bu_n}^{r_j}$, the probability that a region pair $\bu_n$ is associated with the predicate $r_j$, where $r_j \in R$ and $R=\{r_1, r_2,\cdots,r_M\}$ include frequent predicates obtained from the captions in training data. The feed-forward network of  Figure~\ref{fig:vrd}(b) will be trained in weakly supervised training.

\paragraph{Weakly Supervised Multi-Instance Training} As shown in Figure~\ref{fig:vrd}(c), during training, one object pair $t = (o_1, o_2)$, e.g., (\textit{women}, \textit{hat}), can correspond to multiple pairs of object regions: the four women-hat combinations between the two women and two hats. To make our description clearer, we refer to $t = (o_1, o_2)$ as an \textit{object pair}, and the four women-hat pairs in the image as \textit{object region pairs}. Accordingly, for a triple we extracted $t = (o_1, r, o_2), r \in R$, e.g., (\textit{woman}, \textit{in}, \textit{hat}), the predicate $r$ (i.e., \textit{in}) can be associated with multiple \textit{object region pairs} (here, (\textit{w0}, \textit{h0}), (\textit{w0}, \textit{h1}), (\textit{w1}, \textit{h0}), and (\textit{w1}, \textit{h1})). 

To predict predicates over object region pairs, we propose to use Multi-Instance Learning \cite{fang2015captions} as our weakly supervised learning approach. Multi-Instance Learning receives a set of labeled bags, each bag containing a set of instances. A bag would be labeled \textit{negative} if all the instances in it are negative. On the other hand, a bag is labeled \textit{positive} if there is at least one positive instance in the bag.

In our problem, an instance is a region pair. Therefore for a candidate predicate $r \in R$ (e.g., \textit{in}), we use $\mathcal{N}_{r}$ to denote the object region pairs corresponding to predicate $r$. If $r$ appears in the caption $\bm{S}$,  $\mathcal{N}_{r}$ would be a positive bag. We use $\mathcal{N}  \setminus \mathcal{N}_{r}$ to denote the negative bag for ${r}$. When $r$ is not contained in the caption, the entire $\mathcal{N}$ would be the negative bag (the last row of Figure~\ref{fig:vrd}(c)). The probability of a bag $b$ having the predicate $r_j$ is measured with ``noisy-OR":
\begin{equation}
    p_{b}^{r_j} = 1 - \prod_{n \in b} (1 - p_{\bu_n}^{r_j})
\end{equation}
where $p_{\bu_n}^{r_j}$ has been introduced above. We adopt the cross-entropy loss on the basis of all predicate probabilities over bags, given an image $\bm{I}$ and caption $\bm{S}$:
\begin{dmath}
    L(\bm{I}) \!= \!-\!\sum_{j=1}^M \left[\mathbbm{1}_{(r_j \in \bm{S})} (\log p_{\mathcal{N}_{r_j}}^{r_j} \!+\!  \log (1\!-\!p_{\mathcal{N} \setminus \mathcal{N}_{r_j}}^{r_j}))   
         \!+\! \mathbbm{1}_{(r_j \notin \bm{S})}( \log (1\!-\!p_{\mathcal{N}}^{r_j}))\right]
\label{eq:classifier_loss}
\end{dmath}
where the indicator function
$\mathbbm{1}_{condition} = 1$ if the condition is true, otherwise $\mathbbm{1}_{condition} = 0$. 

\label{sec:predicate}

\paragraph{Constructing the Graphs} Once obtaining the trained module, we can build a CGVRG graph $\mathcal{G} = (\mathcal{V}, \mathcal{E})$ for a given image $\bm{I}$, where the node set $\mathcal{V}$ includes two types of nodes: object nodes and predicate nodes. We denote $o_i$ as the $i^{th}$ object node and $r_{ij}$ as a predicate node that connects $o_i$ and $o_j$ (refer to Figure~\ref{fig:weakness} or the middle part of Figure~\ref{fig:framework}). The edges in $\mathcal{E}$ are added based on triples; i.e., $(o_i, r_{ij}, o_j)$ will assign two directed edges from node $o_i$ to $r_{ij}$ and from $r_{ij}$ to $o_j$, respectively.

Note that due to the use of the proposed weakly supervised models, the acquired graphs can now contain predicates that exist in captions but not in the VRD models used in the previous work that does not explicitly consider predicates in captions. We will show in our experiments that this improves captioning quality.

\subsection{Context-Aware CGVRG}
We further enhance CGVRG in the context of both modalities, images and text, using graph convolution networks. We first integrate visual and textual features: the textual features for each node are from a word embedding and the visual features are regional visual representations extracted via RoI pooling from Faster R-CNN.  The specific features $\g_{o_i}, \g_{r_{ij}}$ for object $o_i$ and predicate $r_{ij}$ are shown as follows:
\begin{align}
    \g_{o_i} &= \phi_{o}([\g_{o_i}^t;\g_{o_i}^v]) \\
    \g_{r_{ij}} &= \phi_r(\g_{r_{ij}}^t)
\end{align}
where $\phi_r$ and $\phi_o$ are feed-forward networks using ReLU activation; $\g_{o_i}^t, \g_{r_{ij}}^t$, and $\g_{o_i}^v$ denote textual features of $o_i, r_{ij}$ and visual features of $o_i$, respectively. 

We present the process of encoding $\mathcal{G}$ to produce a new set of context-aware representation $\mathcal{X}$. The representation of predicate $r_{ij}$ and $o_{i}$ are computed as follows:
\begin{align}
   \x_{r_{ij}} = f_{r}([\g_{o_i};\g_{o_j};\g_{r_{ij}}])
\end{align}
\begin{dmath}
    \x_{o_{i}} = \frac{1}{N_{i}}\left[ \sum_{r \in \mathcal{N}_{out}(o_i)} f_{out}([\g_{o_i};\g_{r}]) + \sum_{r \in \mathcal{N}_{in}(o_i)} f_{in}([\g_{o_i};\g_{r}])\right]
\end{dmath}
where $f_{r}, f_{in}, f_{out}$ are feed-forward networks using ReLU activation. $\mathcal{N}_{in}$ and $\mathcal{N}_{out}$ denote the adjacent nodes with $o_i$ as head and tail, respectively. $N_i$ is the total number of adjacent nodes.

\subsection{Multi-task Caption Generation}
\label{sec:caption}
 Unlike the existing image-captioning models, we further incorporate visual relationships into generation --- we propose multi-task learning for jointly predicting word and tag sequences as each word in a caption will be assigned a tag, i.e., \textit{object}, \textit{predicate}, or \textit{none}. The module takes as input the graph node features from the context-aware CGVRG. The output of the generation module is hence the sequence of words $\y = \{y_1,\cdots,y_T\}$ as well as the tags $\bz = \{z_1,\cdots,z_T\}$. Two different approaches are leveraged to train the two tasks jointly. 
 
 The bottom LSTM is used to align a textual state to graph node representations:

\begin{align}
\h_t^{1} &= \LSTM(\h_{t-1}^{1}, [\h_{t-1}^{2}; \overline{\x}; \e_{w_{t}}]) 
\end{align}
where $\LSTM$ means one step of recurrent unit computation via $\LSTM$; $\overline{\x}$ is the mean-pooled representation of all nodes in the graph; $\h_{t-1}^{1}$ and $\h_{t-1}^{2}$ denote hidden states of bottom and top LSTM in time step $t\!-\!1$, respectively; $\e$ is the word embedding table.  

The state $h_t^{1}$ is then used as a query to attend over graph node features $\{\x_{o}\}$ and $ \{\x_{r}\}$ separately to get attended features $\hat{\x}_t^{r}$ and $\hat{\x}_t^{o}$:
\begin{align}
\hat{\x}_t^{r} &= \ATT(\h_t^{1}, \{\x_{r}\})\\
\hat{\x}_t^{o} &= \ATT(\h_t^{1}, \{\x_{o}\})
\end{align}
where $\ATT$ is a soft-attention operation between a query and graph node features. 

The top LSTM works as a language model decoder, in which the hidden state $\h_0^{2}$ is initialized with the mean-pooled semantic representation of all detected predicates $\{r\}$. In time step $t$, the input consists of the output from the bottom LSTM layer $h_{t}^{1}$ and attended graph features $\hat{\x}_t^{r}$,  $\hat{\x}_t^{o}$:
\begin{align}
\h_t^{2} &= \LSTM(\h_{t-1}^{2}, [\h_{t}^{1}; \hat{\x}_t^{o};\hat{\x}_t^{r}])
\end{align}

\subsubsection{Multi-task Learning}
\begin{figure}[htbp]
  \centering
  \includegraphics[width=0.45\textwidth]{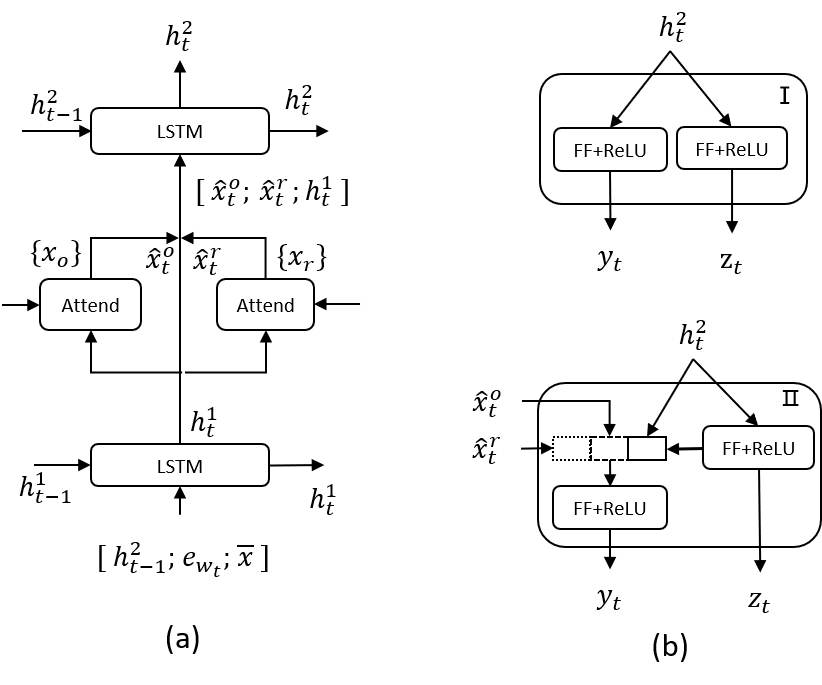}%
  \caption{An overview of multi-task caption generation module. Subfigure (a) is a two-layer LSTM; Subfigure (b) depicts two different types of multi-task block.}\label{fig:att-multi}
\end{figure}

We propose two different blocks to perform the two tasks jointly, as shown in Figure~\ref{fig:att-multi}. In each step, a multi-task learning block deals with task $s_1$ as predicting a tag $z_{t}$ and task $s_2$ as predicting a word $y_{t}$. Specifically
\textbf{MT-\Rmnum{1}} treats the two tasks independent of each other:
\begin{align}
p(z_{t}|y_{< t}, \bm{I}) &= \softmax (f_z(\h_t^2))
\\
p(y_{t}|y_{< t}, \bm{I}) &= \softmax (f_y(\h_t^2))
\label{eq:block1}
\end{align}
where $f_z$ and $f_y$ are feed-forward networks with ReLU activation. Inspired by the adaptive attention mechanism~\cite{lu2017knowing}, \textbf{MT-\Rmnum{2}} further exploits the probability from $p(z_{t}|y_{< t}, \bm{I})$ to integrate the representation of current hidden state $h_t^2$ and attended features from graph $\hat{\x}_t^r,\  \hat{\x}_t^o$:
\begin{align}
p(y_{t}|y_{< t},\bm{I}) &= \softmax (f_y(\hat{\h}_t^2)),
\\
\hat{\h}_t^2 &= \h_t^2p_{na} + \hat{\x}_t^{r}p_{r} + \hat{\x}_t^{o}p_{o}
\\
p(z_{t}|y_{< t},\bm{I}) &= \softmax (f_z(\h_t^2))
\label{eq:block2}
\end{align}
where $p_{na}, p_{r},p_{o}$ denote the probabilities of tag $z_{t}$ being  ``none", ``predicate", and ``object", respectively.
The multi-task loss function is as follows:
\begin{dmath}
L_{MT}(\bm{I}) =\!-\!\sum_{t=1}^{T}\! \log\! p(y_{t}|y_{< t},\bm{I})\!+\!\gamma\!\log\! p(z_{t}|y_{< t},\bm{I})
\label{eq:multi-task loss}
\end{dmath}
where $\gamma$ is the hyper-parameter to balance the two tasks.

\subsection{Training and Inference}
\label{sec:training_stage}
The overall training process can be broken down into two parts: the CGVRG detection module training period and the caption generator training period; the latter includes cross-entropy optimization and the CIDEr-D optimization. For CGVRG detection module training, the detection module is optimized with the multi-instance learning loss in Equation~\ref{eq:classifier_loss}. For caption generator training, the model is first optimized with the cross-entropy loss in Equation~\ref{eq:multi-task loss}, and then we directly optimize the model with
the expected sentence-level reward (CIDEr-D in this work) shown in Equation~\ref{eq:rl_loss} by self critical sequence learning~\cite{rennie2017self}.

In the inference stage, given an image, the CGVRG detection module obtains a graph upon them. The graph convolution network encodes graphs to obtain the context aware multi-modal representations. Then graph object/predicate node features are further provided to the multi-task caption generation module to generate sequences with beam search.

\section{Experiments}

\begin{table*}
\setlength{\tabcolsep}{2.5pt}\small
\centering
\begin{tabular}{l|cccccc|cccccc} \toprule
& \multicolumn{6}{c}{Cross entropy} & \multicolumn{6}{c}{CIDEr-D optimization} \\
\cmidrule{2-13}
& B1 & B4 & ME & RG & CD & SP & B1 & B4 & ME & RG & CD & SP  \\
\midrule

SCST & - & 31.3 & 26.0 & 54.3 & 101.3 & - & - & 33.3 & 26.3 & 55.3 & 111.4 & - \\
LSTM-A & 75.4 & 35.2 & 26.9 & 55.8 & 108.8 & 20.0 & 78.6 & 35.5 & 27.3 & 56.8 & 118.3 & 20.8 \\
Up-Down (Baseline) & 77.2 & 36.2 & 27.0 & 56.4 & 113.5 & 20.3  & 79.8 & 36.3 & 27.7 & 56.9 & 120.1 & 21.4 \\
StackCap & 76.2 & 35.2 & 26.5 & - & 109.1 & - & 78.6 & 36.1 & 27.4 & - & 120.4 & -\\
CAVP & - & - & - & - & - & - & - & 38.6 & 28.3 & 58.5 & 126.3 & 21.6\\
GCN-LSTM & 77.3 & 36.8 & 27.9 & 57.0 & 116.3 & 20.9 & 80.5 & 38.2 & 28.5 & 58.3 & 127.6 & 22.0 \\
VSUA & - & - & - & - & - & - & - & 38.4 & 28.5 & 58.4 & 128.6 & 22.0\\
SGAE & 77.6 & 36.9 & 27.7 & 57.2 & 116.7 & 20.9 & 80.8 & 38.4 & 28.4 & 58.6 & 127.8 & 22.1\\
\midrule
This Work (MT-\Rmnum{1})  & \textbf{78.1} & \textbf{38.4} & \textbf{28.2} & \textbf{58.0} & \textbf{119.0} & 21.1 & \textbf{80.8} & \textbf{38.9} & \textbf{28.8} & \textbf{58.7} & \textbf{129.6} & 22.3 \\ 
This Work (MT-\Rmnum{2}) & 77.9 & 38.0 & 28.1 & 57.6 & 117.8 & \textbf{21.3} & 80.5 & 38.6 & 28.7 & 58.4 & 128.7 & \textbf{22.4} \\ 
\bottomrule
\end{tabular}
\caption{Single-model performances on the MSCOCO dataset (Karpathy split) in both cross-entropy and RL training period. B1, B4, ME, RG, CD, and SP denote BLEU-1, BLEU-4, METEOR, ROUGE, CIDEr-D and SPICE, respectively.}
\label{tab:coco-offline}
\end{table*}

\subsection{Datasets and Experiment Setup}
\paragraph{MSCOCO} We perform extensive experiments on the MSCOCO benchmark~\cite{lin2014microsoft}. The Karpathy split~\cite{karpathy2015deep} is adopted for our model selection and offline testing, which contains 113K
training images, 5K validation images 
and 5K testing images. As for the online test server, the result is trained on the entire training and validation set (123K images). To evaluate the generated captions, we employ standard evaluation metrics: SPICE~\cite{anderson2016spice},
CIDEr-D~\cite{vedantam2015cider}, METEOR~\cite{denkowski2014meteor}, ROUGE-L~\cite{lin2004rouge}, and BLEU ~\cite{papineni2002bleu}.

\paragraph{Visual Genome} We use the Visual Genome~\cite{krishna2017visual} dataset to pre-train our object detection model. The dataset includes 108K images. To pre-train the object detection model with Faster R-CNN, we strictly follow the setting in~\cite{anderson2018bottom}, taking
98K/5K/5K for training, validation, and testing, respectively. The split is carefully selected to avoid contamination of the MSCOCO validation and testing sets, since nearly 51K Visual Genome images are also included in the MSCOCO dataset. 

\paragraph{Implementation Details}
We use Faster R-CNN~\cite{ren2015faster} to identify and localize instances of objects. The object detection phase consists of two modules. The first module proposes object regions using a deep CNN, i.e., ResNet-101~\cite{he2016deep}. The second module extracts feature maps using region-of-interest pooling for each box proposals. Practically, we take the final output of the ResNet-101 and perform non-maximum suppression for each object class with an IoU threshold. As a result, we obtain a set of image regions, $V = \{v_1, v_2, \cdots, v_n\}$, where $n \in [10, 100]$ varies with input images and confidence thresholds. Each region is represented as a 2,048-dimensional vector obtained from the pool5 layer after the RoI pooling. We then apply a feed-forward network with a 1000-dimensional output layer for predicates classification. The network of the same size is also used for feature projection ($\phi_o, \phi_i$) and GCN ($f_r, f_{in}, f_{out}$). In the decoder LSTM, the word embedding dimension is set to be 1,000 and the hidden unit dimension in the top-layer and bottom-layer LSTM is set to be 1,000 and 512, respectively. The trade-off parameter $\gamma$ in multi-task learning is 0.15. The whole
system is trained with the Adam optimizer. We set the initial learning rate to be
0.0005 and mini-batch size to be 100. The maximum number of training epochs is 30 for Cross-entropy and CIDEr-D optimization respectively. For sequence generation in the inference stage, we adopt the beam
search strategy and set the beam size to be 3.

We construct object and predicate categories for VRD training. Similar to~\cite{lu2018neural}, we manually expand the original 80 object categories to 413 fine-grained categories by utilizing a list of caption tokens. For example, the object category \textit{``person"} is expanded to a list of fine-grained categories $[\textit{``boy”}, \textit{``man”}, \cdots]$. Then for all extracted triples that have both objects appearing in the 413 category list, we select the 200 most frequent predicates as our predicate categories. 


\subsection{Quantitative Analysis}
\begin{table}
\setlength{\tabcolsep}{2.5pt}\small
\centering
\begin{tabular}{l|cccccccccc} \toprule
& 
\multicolumn{2}{c}{B4} &
\multicolumn{2}{c}{ME} &
\multicolumn{2}{c}{RG} &
\multicolumn{2}{c}{CD}
\\
\cmidrule{2-9}
& c5 & c40 & c5 & c40 & c5 & c40 & c5 & c40 \\
\midrule
GCN-LSTM* 
& \textbf{38.7} & 69.7 & 28.5 & 37.6 & 58.5 & 73.4 & \textbf{125.3} & 126.5\\
VSUA 
& 37.4 & 68.3 & 28.2 & 37.1 & 57.9 & 72.8 & 123.1 & 125.5\\
SGAE 
& 38.5 & 69.7 & 28.2 & 37.2 & 58.6 & 73.6 & 123.8 & 126.5\\
\midrule
Baseline
& 36.9 & 68.5 & 27.6 & 36.7 & 57.1 & 72.4 & 117.9 & 120.5 \\
This Work 
& 38.6 & \textbf{70.1} & \textbf{28.6} & \textbf{37.8} & \textbf{58.8} & \textbf{74.5} & 125.1 & \textbf{126.7} \\ 

\bottomrule
\end{tabular}
\caption{The performance on COCO online test server of various methods that incorporate visual relationships. * denotes that their training batch size and epochs are far beyond average setting in ~\cite{anderson2018bottom,yang2019auto}.}
\label{tab:coco-online}
\end{table}

\paragraph{Model Comparison} We compare our models with the following state-of-the-art models: (1) SCST~\cite{rennie2017self} employs an improved policy gradient algorithm by utilizing its own inference output to normalize the rewards; (2) LSTM-A~\cite{yao2017boosting} integrates the detected image attributes into the CNN-plus-RNN image captioning framework; (3) Up-Down~\cite{anderson2018bottom} uses both a bottom-up and top-down attention mechanism to focus more on salient object regions; (4) GCN-LSTM~\cite{yao2018exploring} leverages graph convolutional networks over the detected objects and relations; (5) CAVP~\cite{liu2018context} proposes a context-aware policy network by accounting for visual attentions as context for generation; (6) VSUA~\cite{guo2019aligning} exploits the alignment between words and different categories of graph nodes; (7) SAGE~\cite{yang2019auto} utilizes an additional graph encoder to incorporate language inductive bias into the encoder-decoder framework.

Our baseline is built on Up-Down~\cite{anderson2018bottom}. We propose two variants of final models using different multi-task blocks, namely MT-\Rmnum{1} and MT-\Rmnum{2} shown in Fig~\ref{fig:att-multi}(b). We conduct extensive comparisons on the dataset with the above state-of-the-art techniques. We also perform detailed analysis to demonstrate the impact of different components of our framework.



Table~\ref{tab:coco-offline} lists the results of various single models on
the MSCOCO Karpathy split. Our model outperforms the baseline model significantly, with CIDEr-D scores being improved from 113.5 to 119.0 and 120.1 to 129.6 in the cross-entropy and CIDEr-D optimization period, respectively. 
In addition, the model with MT-\Rmnum{2} shows an advantage over that with MT-\Rmnum{1} on SPICE, which implies that the proposed adaptive visual attention mechanism works in multi-task block \Rmnum{2}. 


Table~\ref{tab:coco-online} compares our model with three models that also incorporate VRG, plus the baseline model, on the MSCOCO online test server. Our model improves significantly from the baseline (from 120.5 to 126.7 in CIDEr-D) and has achieved the best results across all evaluation metrics on c40 (40 reference captions). 


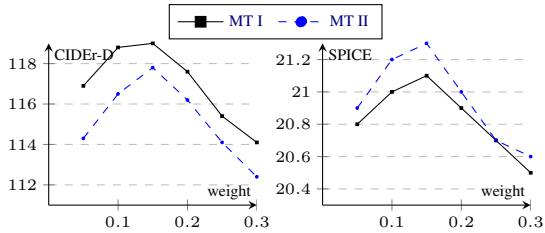
\begin{figure}[!t] \tiny 
  \centering
  \pgfplotsset{width=0.27\textwidth}
  \ref{aa}\\
\begin{tikzpicture}
    \begin{axis}[
    xlabel={weight},
   ylabel={CIDEr-D},
   xmin=0,
        ymin=111,
    mark size=0.5pt,
    ymajorgrids=true,
    grid style=dashed,
    legend columns=-1,
    legend entries={MT \Rmnum{1}, MT \Rmnum{2}},
    legend style={/tikz/every even column/.append style={column sep=0.13cm}},
    legend to name=aa,
    ]
    \addplot [black, mark=square*] table [x index=0, y index=3] {multi-weight_new.txt};
    \addplot [blue, dashed, mark=*] table [x index=0, y index=1] {multi-weight_new.txt};
    \end{axis}
\end{tikzpicture}
\hspace{-0.8em}
\begin{tikzpicture}
    \begin{axis}[
    xlabel={weight},
    ylabel={SPICE},
    xmin=0,
        ymin=20.3,
    mark size=0.5pt,
    ymajorgrids=true,
    grid style=dashed,
    ]
    \addplot [black, mark=square*] table [x index=0, y index=4] {multi-weight_new.txt};
    \addplot [blue, dashed, mark=*] table [x index=0, y index=2] {multi-weight_new.txt};
    \end{axis}
\end{tikzpicture}
\caption{Test results (cross-entropy optimization) on various $\gamma$.}
\label{fig:multitask-weight}
\end{figure}

\begin{figure}[!t]
  \centering
  \includegraphics[width=0.45\textwidth]{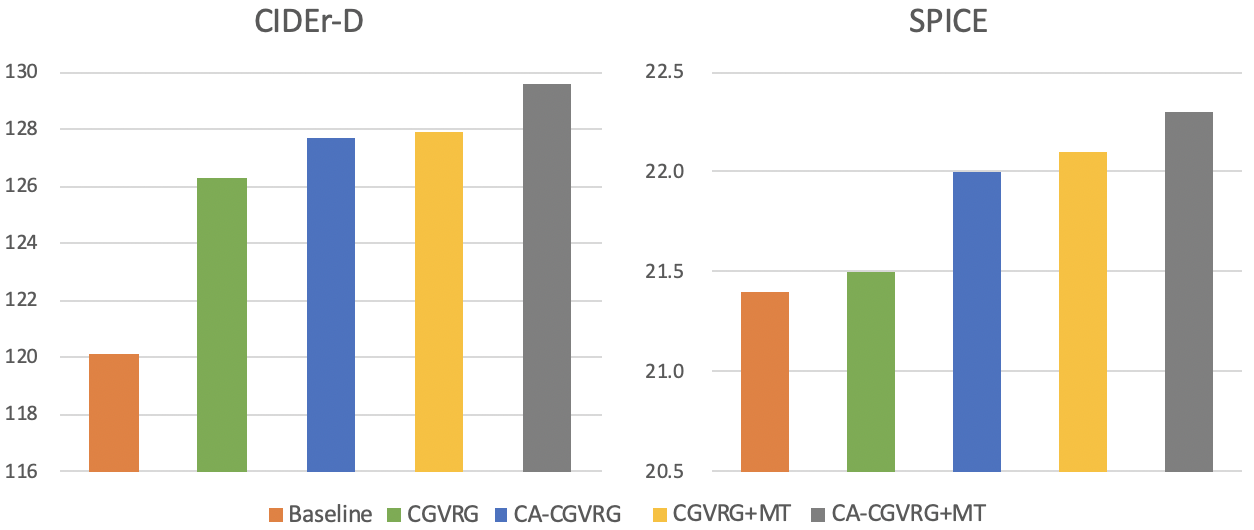}
  \caption{Ablation results (CIDEr-D optimization). }\label{fig:ablation}
\end{figure}

\begin{figure*}[htbp]
  \centering
  \includegraphics[width=1.0\textwidth]{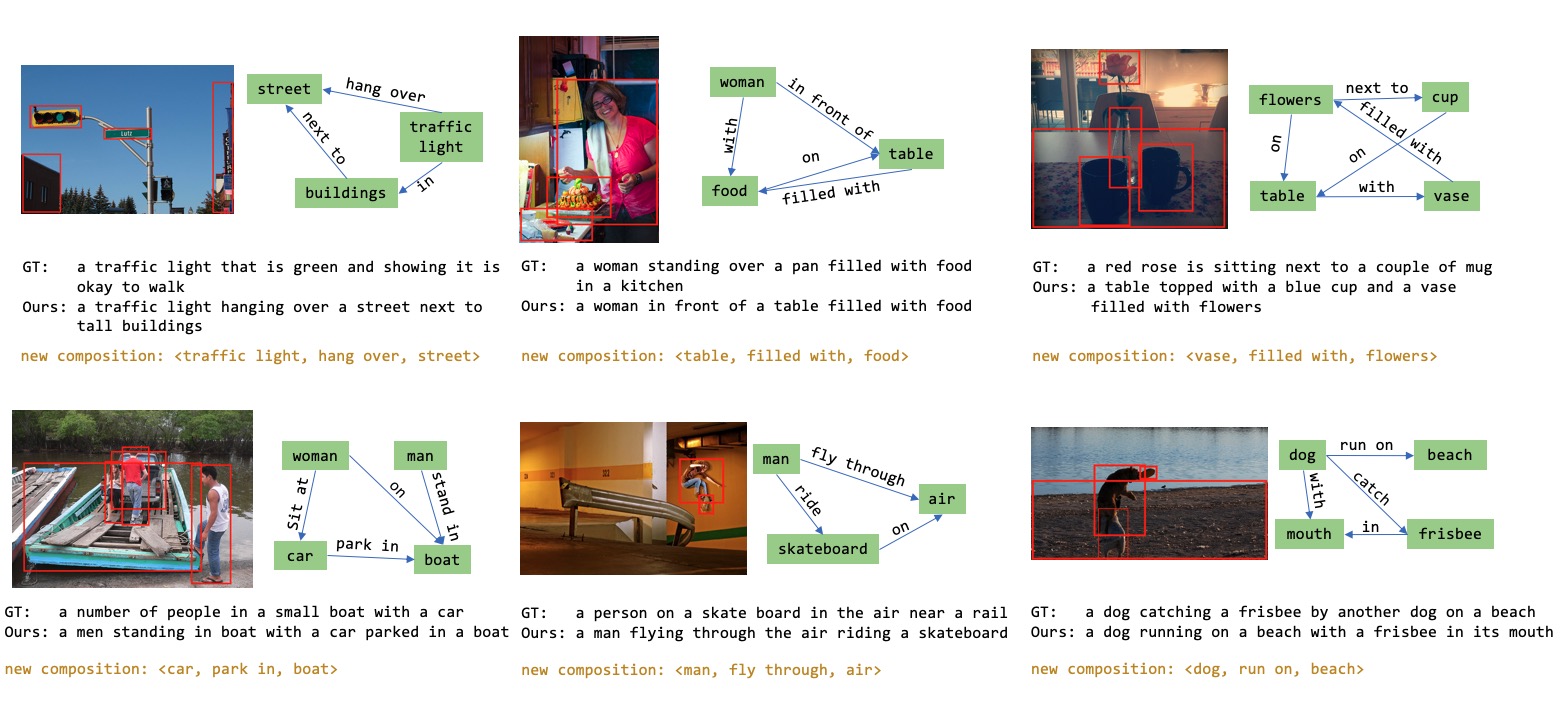}
  \caption{Several image captioning examples generated by our model. }\label{fig:example-case}
\end{figure*}
\begin{figure}[htbp]
  \centering
  \includegraphics[width=0.5\textwidth]{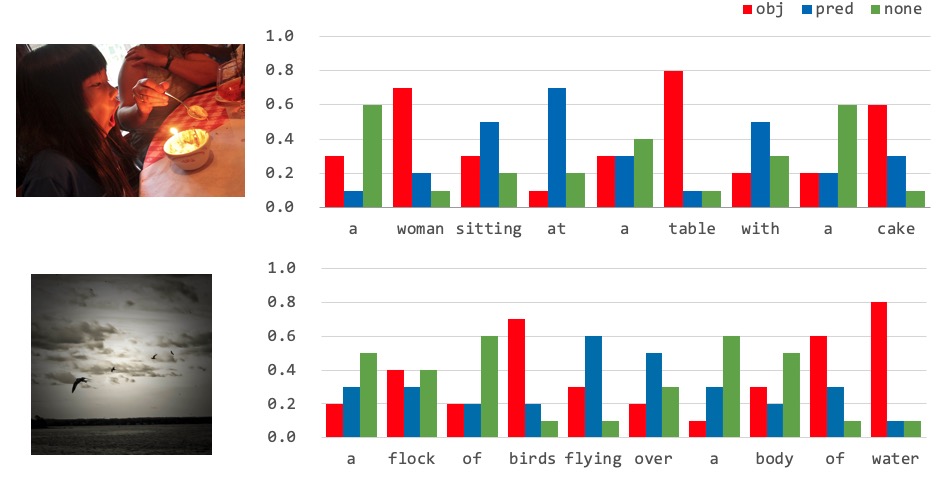}
  \caption{Examples of generated word and tag sequences. }\label{fig:example2-mt}
\end{figure}

Figure~\ref{fig:multitask-weight} shows the effect of taking different weights $\gamma$ in the multi-task loss item (Equation~\ref{eq:multi-task loss}). The results indicate that the weight around 0.15 yields the best performance in both multi-task blocks. Meanwhile, Figure~\ref{fig:ablation} shows the ablation analysis by removing the multi-task caption generation and graph convolution operation, respectively, to check the effect of these components. The results show that both the graph convolution operation and multi-task learning help improve the quality of the generated captions.

Note that the code of our paper has been made publicly available in the webpage provided in the abstract.

\paragraph{Human evaluation} 
We performed human evaluation with three non-author human subjects, using a five-level Likert scale. For each image and each pair of systems in comparison (MT-I vs. Up-Down, MT-I vs. GCN-LSTM, and MT-I vs. SGAE), we show the captions generated by the two systems to the human subjects. We ask each subject if the first caption sentence is: significantly better ($2$), better ($1$), equal ($0$), worse ($-1$), or significantly worse ($-2$), compared to the second. 

Following~\cite{zhao2019informative}, we obtain the subjects' ratings for fidelity (the first caption is superior in terms of making less mistakes?), informativeness (the first caption provides more informative and detailed description?), and fluency (the first caption is more fluent?). For each question asked for an image, we calculate the average of the three subjects’ scores.
For each pair of models in comparison, we randomly sampled 50 images from the Karpathy testset.

\begin{itemize}
    \item MT-\Rmnum{1} vs. Up-Down:
For fidelity, MT-I is better or significantly better on 44\% images (where the average of the three human subjects’ scores is larger than $0.5$), equal to Up-Down on 46\% images (the average is in range $[-0.5,0.5]$), and worse or significantly worse on 10\% images (average is less than $-0.5$). For informativeness, MT-I is better or significantly better on 60\% images, equal on 34\%, and worse or significantly worse on 6\%. For fluency, the numbers are 18\%, 72\%, and 10\%. 
    \item MT-\Rmnum{1} vs. GCN-LSTM:
For fidelity, MT-I is better or significantly better on 40\% images, equal to GCN-LSTM on 52\%, and worse or significantly worse on 8\%. For informativeness, the numbers are 32\%, 50\%, and 18\%, respectively. For fluency, the numbers are 12\%, 76\%, and 12\%.
    \item MT-\Rmnum{1} vs. SGAE:
For fidelity, MT-I is better or significantly better on 36\% images, equal to SGAE on 56\%, and worse or significantly worse on 8\%. For informativeness, the numbers are 30\%, 48\%, and 22\%, respectively. For fluency, the numbers are 6\%, 90\%, and 4\%. 
\end{itemize}

\subsection{Qualitative Analysis}
Figure~\ref{fig:example-case} shows several specific examples, each including an image, a detected caption guided visual relationship graph, a ground truth sentence, a generated word sequence, and a learned visual relationship composition. We can see that the proposed model generates more accurate captions coherent to the visual relationship detected in the image. Consider the upper middle demo as an example; our model extracts a visual relationship graph covering the critical predicates ``filled with'' and ``in front of'' for understanding the image, thus producing a comprehensive description. In addition, we observe that the model generates the triple $(table, filled\ with, food)$, which is a new composition that has not appeared in the training set.

Figure~\ref{fig:example2-mt} visualizes the effect of our tag sequence generation process. Specifically, we visualize the tag probabilities of the ``object'', ``predicate'', and ``none'' category in each generation step. Our model successfully learns to distinguish the correct category for each time step, which is in consistent with the tag of the predicted word.  For example, for the generated words ``flying over", the probability for the ``predicate" category is the highest, which is also true for words like ``bird" and ``water".

\section{Conclusions}

This paper presents a novel image captioning architecture that constructs caption-guided visual relationship graphs to introduce beneficial inductive bias to better utilize captions. The representation is further enhanced with text and visual features of neighbouring nodes. During generation, the network is regularized to take into account explicit object/predicate constraints with multi-task learning. Extensive experiments are performed on the MSCOCO dataset, showing that the proposed framework significantly outperforms the baselines, resulting in the state-of-the-art performance under various evaluation metrics. In the near future we plan to extend the proposed approach to several other language-vision modeling tasks.

\section*{Acknowledgements}
We would like to thank the anonymous reviewers for their valuable comments. This research of the first and last author is supported by the Natural Sciences and Engineering Research Council of Canada (NSERC).

\bibliography{acl2020}

\begin{thebibliography}{37}
\expandafter\ifx\csname natexlab\endcsname\relax\def\natexlab#1{#1}\fi

\bibitem[{Anderson et~al.(2016)Anderson, Fernando, Johnson, and
  Gould}]{anderson2016spice}
Peter Anderson, Basura Fernando, Mark Johnson, and Stephen Gould. 2016.
\newblock Spice: Semantic propositional image caption evaluation.
\newblock In \emph{ECCV}, pages 382--398. Springer.

\bibitem[{Anderson et~al.(2018)Anderson, He, Buehler, Teney, Johnson, Gould,
  and Zhang}]{anderson2018bottom}
Peter Anderson, Xiaodong He, Chris Buehler, Damien Teney, Mark Johnson, Stephen
  Gould, and Lei Zhang. 2018.
\newblock Bottom-up and top-down attention for image captioning and visual
  question answering.
\newblock In \emph{CVPR}, pages 6077--6086.

\bibitem[{Chen et~al.(2017)Chen, Zhang, Xiao, Nie, Shao, Liu, and
  Chua}]{chen2017sca}
Long Chen, Hanwang Zhang, Jun Xiao, Liqiang Nie, Jian Shao, Wei Liu, and
  Tat-Seng Chua. 2017.
\newblock Sca-cnn: Spatial and channel-wise attention in convolutional networks
  for image captioning.
\newblock In \emph{CVPR}, pages 5659--5667.

\bibitem[{Chen et~al.(2015)Chen, Fang, Lin, Vedantam, Gupta, Doll{\'a}r, and
  Zitnick}]{chen2015microsoft}
Xinlei Chen, Hao Fang, Tsung-Yi Lin, Ramakrishna Vedantam, Saurabh Gupta, Piotr
  Doll{\'a}r, and C~Lawrence Zitnick. 2015.
\newblock Microsoft coco captions: Data collection and evaluation server.
\newblock \emph{arXiv}.

\bibitem[{Dai et~al.(2017)Dai, Zhang, and Lin}]{dai2017detecting}
Bo~Dai, Yuqi Zhang, and Dahua Lin. 2017.
\newblock Detecting visual relationships with deep relational networks.
\newblock In \emph{CVPR}, pages 3076--3086.

\bibitem[{Denkowski and Lavie(2014)}]{denkowski2014meteor}
Michael Denkowski and Alon Lavie. 2014.
\newblock Meteor universal: Language specific translation evaluation for any
  target language.
\newblock In \emph{Proceedings of the ninth workshop on statistical machine
  translation}, pages 376--380.

\bibitem[{Donahue et~al.(2015)Donahue, Anne~Hendricks, Guadarrama, Rohrbach,
  Venugopalan, Saenko, and Darrell}]{donahue2015long}
Jeffrey Donahue, Lisa Anne~Hendricks, Sergio Guadarrama, Marcus Rohrbach,
  Subhashini Venugopalan, Kate Saenko, and Trevor Darrell. 2015.
\newblock Long-term recurrent convolutional networks for visual recognition and
  description.
\newblock In \emph{CVPR}, pages 2625--2634.

\bibitem[{Fang et~al.(2015)Fang, Gupta, Iandola, Srivastava, Deng, Doll{\'a}r,
  Gao, He, Mitchell, Platt et~al.}]{fang2015captions}
Hao Fang, Saurabh Gupta, Forrest Iandola, Rupesh~K Srivastava, Li~Deng, Piotr
  Doll{\'a}r, Jianfeng Gao, Xiaodong He, Margaret Mitchell, John~C Platt,
  et~al. 2015.
\newblock From captions to visual concepts and back.
\newblock In \emph{CVPR}, pages 1473--1482.

\bibitem[{Gan et~al.(2017)Gan, Gan, He, Pu, Tran, Gao, Carin, and
  Deng}]{gan2017semantic}
Zhe Gan, Chuang Gan, Xiaodong He, Yunchen Pu, Kenneth Tran, Jianfeng Gao,
  Lawrence Carin, and Li~Deng. 2017.
\newblock Semantic compositional networks for visual captioning.
\newblock In \emph{CVPR}, pages 5630--5639.

\bibitem[{Guo et~al.(2019)Guo, Liu, Tang, Li, Luo, and Lu}]{guo2019aligning}
Longteng Guo, Jing Liu, Jinhui Tang, Jiangwei Li, Wei Luo, and Hanqing Lu.
  2019.
\newblock Aligning linguistic words and visual semantic units for image
  captioning.
\newblock \emph{arXiv}.

\bibitem[{He et~al.(2016)He, Zhang, Ren, and Sun}]{he2016deep}
Kaiming He, Xiangyu Zhang, Shaoqing Ren, and Jian Sun. 2016.
\newblock Deep residual learning for image recognition.
\newblock In \emph{CVPR}, pages 770--778.

\bibitem[{Johnson et~al.(2015)Johnson, Krishna, Stark, Li, Shamma, Bernstein,
  and Fei-Fei}]{johnson2015image}
Justin Johnson, Ranjay Krishna, Michael Stark, Li-Jia Li, David Shamma, Michael
  Bernstein, and Li~Fei-Fei. 2015.
\newblock Image retrieval using scene graphs.
\newblock In \emph{CVPR}, pages 3668--3678.

\bibitem[{Karpathy and Fei-Fei(2015)}]{karpathy2015deep}
Andrej Karpathy and Li~Fei-Fei. 2015.
\newblock Deep visual-semantic alignments for generating image descriptions.
\newblock In \emph{CVPR}, pages 3128--3137.

\bibitem[{Kim et~al.(2019)Kim, Choi, Oh, and Kweon}]{kim2019dense}
Dong-Jin Kim, Jinsoo Choi, Tae-Hyun Oh, and In~So Kweon. 2019.
\newblock Dense relational captioning: Triple-stream networks for
  relationship-based captioning.
\newblock In \emph{Proceedings of the IEEE Conference on Computer Vision and
  Pattern Recognition}, pages 6271--6280.

\bibitem[{Krishna et~al.(2017)Krishna, Zhu, Groth, Johnson, Hata, Kravitz,
  Chen, Kalantidis, Li, Shamma et~al.}]{krishna2017visual}
Ranjay Krishna, Yuke Zhu, Oliver Groth, Justin Johnson, Kenji Hata, Joshua
  Kravitz, Stephanie Chen, Yannis Kalantidis, Li-Jia Li, David~A Shamma, et~al.
  2017.
\newblock Visual genome: Connecting language and vision using crowdsourced
  dense image annotations.
\newblock \emph{International Journal of Computer Vision}, 123(1):32--73.

\bibitem[{Li et~al.(2017)Li, Ouyang, Wang, and Tang}]{li2017vip}
Yikang Li, Wanli Ouyang, Xiaogang Wang, and Xiao'ou Tang. 2017.
\newblock Vip-cnn: Visual phrase guided convolutional neural network.
\newblock In \emph{CVPR}, pages 1347--1356.

\bibitem[{Lin(2004)}]{lin2004rouge}
Chin-Yew Lin. 2004.
\newblock Rouge: A package for automatic evaluation of summaries.
\newblock \emph{Text Summarization Branches Out}.

\bibitem[{Lin et~al.(2014)Lin, Maire, Belongie, Hays, Perona, Ramanan,
  Doll{\'a}r, and Zitnick}]{lin2014microsoft}
Tsung-Yi Lin, Michael Maire, Serge Belongie, James Hays, Pietro Perona, Deva
  Ramanan, Piotr Doll{\'a}r, and C~Lawrence Zitnick. 2014.
\newblock Microsoft coco: Common objects in context.
\newblock In \emph{ECCV}, pages 740--755. Springer.

\bibitem[{Liu et~al.(2018)Liu, Zha, Zhang, Zhang, and Wu}]{liu2018context}
Daqing Liu, Zheng-Jun Zha, Hanwang Zhang, Yongdong Zhang, and Feng Wu. 2018.
\newblock Context-aware visual policy network for sequence-level image
  captioning.
\newblock \emph{arXiv}.

\bibitem[{Lu et~al.(2016)Lu, Krishna, Bernstein, and Fei-Fei}]{lu2016visual}
Cewu Lu, Ranjay Krishna, Michael Bernstein, and Li~Fei-Fei. 2016.
\newblock Visual relationship detection with language priors.
\newblock In \emph{ECCV}, pages 852--869. Springer.

\bibitem[{Lu et~al.(2017)Lu, Xiong, Parikh, and Socher}]{lu2017knowing}
Jiasen Lu, Caiming Xiong, Devi Parikh, and Richard Socher. 2017.
\newblock Knowing when to look: Adaptive attention via a visual sentinel for
  image captioning.
\newblock In \emph{CVPR}, pages 375--383.

\bibitem[{Lu et~al.(2018)Lu, Yang, Batra, and Parikh}]{lu2018neural}
Jiasen Lu, Jianwei Yang, Dhruv Batra, and Devi Parikh. 2018.
\newblock Neural baby talk.
\newblock In \emph{CVPR}, pages 7219--7228.

\bibitem[{Marcheggiani and Titov(2017)}]{marcheggiani2017encoding}
Diego Marcheggiani and Ivan Titov. 2017.
\newblock Encoding sentences with graph convolutional networks for semantic
  role labeling.
\newblock \emph{arXiv}.

\bibitem[{Papineni et~al.(2002)Papineni, Roukos, Ward, and
  Zhu}]{papineni2002bleu}
Kishore Papineni, Salim Roukos, Todd Ward, and Wei-Jing Zhu. 2002.
\newblock Bleu: a method for automatic evaluation of machine translation.
\newblock In \emph{ACL}, pages 311--318.

\bibitem[{Ren et~al.(2015)Ren, He, Girshick, and Sun}]{ren2015faster}
Shaoqing Ren, Kaiming He, Ross Girshick, and Jian Sun. 2015.
\newblock Faster r-cnn: Towards real-time object detection with region proposal
  networks.
\newblock In \emph{NIPS}, pages 91--99.

\bibitem[{Rennie et~al.(2017)Rennie, Marcheret, Mroueh, Ross, and
  Goel}]{rennie2017self}
Steven~J Rennie, Etienne Marcheret, Youssef Mroueh, Jerret Ross, and Vaibhava
  Goel. 2017.
\newblock Self-critical sequence training for image captioning.
\newblock In \emph{CVPR}, pages 7008--7024.

\bibitem[{Schuster et~al.(2015)Schuster, Krishna, Chang, Fei-Fei, and
  Manning}]{schuster2015generating}
Sebastian Schuster, Ranjay Krishna, Angel Chang, Li~Fei-Fei, and Christopher~D.
  Manning. 2015.
\newblock Generating semantically precise scene graphs from textual
  descriptions for improved image retrieval.
\newblock In \emph{Workshop on Vision and Language (VL15)}, Lisbon, Portugal.
  ACL.

\bibitem[{Vedantam et~al.(2015)Vedantam, Lawrence~Zitnick, and
  Parikh}]{vedantam2015cider}
Ramakrishna Vedantam, C~Lawrence~Zitnick, and Devi Parikh. 2015.
\newblock Cider: Consensus-based image description evaluation.
\newblock In \emph{CVPR}, pages 4566--4575.

\bibitem[{Vinyals et~al.(2015)Vinyals, Toshev, Bengio, and
  Erhan}]{vinyals2015show}
Oriol Vinyals, Alexander Toshev, Samy Bengio, and Dumitru Erhan. 2015.
\newblock Show and tell: A neural image caption generator.
\newblock In \emph{CVPR}, pages 3156--3164.

\bibitem[{Wu et~al.(2016)Wu, Shen, Liu, Dick, and Van Den~Hengel}]{wu2016value}
Qi~Wu, Chunhua Shen, Lingqiao Liu, Anthony Dick, and Anton Van Den~Hengel.
  2016.
\newblock What value do explicit high level concepts have in vision to language
  problems?
\newblock In \emph{CVPR}, pages 203--212.

\bibitem[{Xu et~al.(2015)Xu, Ba, Kiros, Cho, Courville, Salakhudinov, Zemel,
  and Bengio}]{xu2015show}
Kelvin Xu, Jimmy Ba, Ryan Kiros, Kyunghyun Cho, Aaron Courville, Ruslan
  Salakhudinov, Rich Zemel, and Yoshua Bengio. 2015.
\newblock Show, attend and tell: Neural image caption generation with visual
  attention.
\newblock In \emph{ICML}, pages 2048--2057.

\bibitem[{Yang et~al.(2019)Yang, Tang, Zhang, and Cai}]{yang2019auto}
Xu~Yang, Kaihua Tang, Hanwang Zhang, and Jianfei Cai. 2019.
\newblock Auto-encoding scene graphs for image captioning.
\newblock In \emph{CVPR}, pages 10685--10694.

\bibitem[{Yao et~al.(2018)Yao, Pan, Li, and Mei}]{yao2018exploring}
Ting Yao, Yingwei Pan, Yehao Li, and Tao Mei. 2018.
\newblock Exploring visual relationship for image captioning.
\newblock In \emph{ECCV}, pages 684--699.

\bibitem[{Yao et~al.(2017)Yao, Pan, Li, Qiu, and Mei}]{yao2017boosting}
Ting Yao, Yingwei Pan, Yehao Li, Zhaofan Qiu, and Tao Mei. 2017.
\newblock Boosting image captioning with attributes.
\newblock In \emph{ICCV}, pages 4894--4902.

\bibitem[{You et~al.(2016)You, Jin, Wang, Fang, and Luo}]{you2016image}
Quanzeng You, Hailin Jin, Zhaowen Wang, Chen Fang, and Jiebo Luo. 2016.
\newblock Image captioning with semantic attention.
\newblock In \emph{CVPR}, pages 4651--4659.

\bibitem[{Zhang et~al.(2017)Zhang, Kyaw, Chang, and Chua}]{zhang2017visual}
Hanwang Zhang, Zawlin Kyaw, Shih-Fu Chang, and Tat-Seng Chua. 2017.
\newblock Visual translation embedding network for visual relation detection.
\newblock In \emph{CVPR}, pages 5532--5540.

\bibitem[{Zhao et~al.(2019)Zhao, Sharma, Levinboim, and
  Soricut}]{zhao2019informative}
Sanqiang Zhao, Piyush Sharma, Tomer Levinboim, and Radu Soricut. 2019.
\newblock Informative image captioning with external sources of information.
\newblock \emph{arXiv preprint arXiv:1906.08876}.

\end{thebibliography}
\bibliographystyle{acl_natbib}
\end{document}